\documentclass[runningheads]{llncs}
\usepackage[T1]{fontenc}
\usepackage{graphicx,verbatim}
\usepackage{booktabs}
\usepackage{multirow}
\usepackage{graphicx}
\usepackage{orcidlink}
\usepackage{xcolor}
\usepackage{subcaption}
\usepackage{amsmath}
\usepackage{color}

\begin{document}
\title{MedSymmFlow: Bridging Generative Modeling and Classification in Medical Imaging through Symmetrical Flow Matching}
\titlerunning{MedSymmFlow: Bridging Generative Modeling and Classification}
%
\author{Francisco Caetano~\orcidlink{0000-0002-6069-6084}\inst{1} \and
Lemar Abdi~\orcidlink{0009-0003-9301-1613}\inst{1} \and
Christiaan Viviers~\orcidlink{0000-0001-6455-0288}\inst{1} \and
Amaan Valiuddin~\orcidlink{0009-0005-2856-5841}\inst{1} \and
Fons van der Sommen~\orcidlink{0000-0002-3593-2356}\inst{1}}
\authorrunning{F. Caetano et al.}
%
\institute{Eindhoven University of Technology, 5612 AE, Eindhoven, Netherlands
        \email{f.t.de.espirito.santo.e.caetano@tue.nl}}



\maketitle 
\begin{abstract}
Reliable medical image classification requires accurate predictions and well-calibrated uncertainty estimates, especially in high-stakes clinical settings. This work presents MedSymmFlow, a generative-discriminative hybrid model built on Symmetrical Flow Matching, designed to unify classification, generation, and uncertainty quantification in medical imaging. MedSymmFlow leverages a latent-space formulation that scales to high-resolution inputs and introduces a semantic mask conditioning mechanism to enhance diagnostic relevance. Unlike standard discriminative models, it naturally estimates uncertainty through its generative sampling process. The model is evaluated on four MedMNIST datasets, covering a range of modalities and pathologies. The results show that MedSymmFlow matches or exceeds the performance of established baselines in classification accuracy and AUC, while also delivering reliable uncertainty estimates validated by performance improvements under selective prediction. The code is available at \href{https://github.com/caetas/MedSymmFlow}{github.com/caetas/MedSymmFlow}.
\end{abstract}

\section{Introduction}

Comprehending semantic content within medical images is paramount for various clinical applications. Tasks such as disease classification~\cite{wang2017chestx,tschandl2018ham10000,kermany2018identifying,acevedo2020dataset} and segmentation~\cite{wu2024medsegdiff,ma2024segment,vezakis2024effisegnet} enable detailed analysis and structure of medical data, facilitating diagnosis and treatment planning. In addition, generative modeling has immense potential for augmenting datasets and synthesizing realistic medical imagery. Ideally, a unified framework would bridge these capabilities, allowing models to both interpret and generate medical images in a mutually beneficial manner. It is conceivable that a deeper understanding and disentanglement of pathological structures would lead to more clinically relevant and realistic synthetic data. In contrast, robust generative capabilities might aid in learning more expressive representations of medical images, capturing subtle disease patterns and contextual information. This inherent reciprocity suggests that advancements in one area could naturally enhance the other, motivating the development of models that integrate both understanding and synthesis within a cohesive framework.

Current deep learning approaches in medical imaging often address these tasks in isolation. Discriminative classifiers extract features to predict diagnostic labels~\cite{manzari2023medvit}, while segmentation models learn to delineate regions of interest~\cite{tang2022self}. 
Although recent efforts have explored the use of diffusion models for classification~\cite{favero2025conditional} and segmentation~\cite{wu2024medsegdiff} in medical imaging, these adaptations can introduce limitations. For instance, diffusion-based classification might involve iterative sampling, potentially increasing inference time, and segmentation frameworks may primarily focus on mask generation without a direct pathway back to realistic image synthesis.

Inspired by recent progress in unifying generative and discriminative modeling within natural image domains, this work introduces MedSymmFlow, a novel framework that extends Symmetrical Flow Matching~\cite{caetano2025symmetrical} to medical imaging. MedSymmFlow leverages the strengths of generative modeling to enable accurate classification while inherently providing explainability and uncertainty estimates. Specifically, we (a)~propose a semantic RGB mask-label strategy that enhances multiclass conditioning, overcoming the limitations of grayscale conditioning; (b)~develop a latent-space formulation of Symmetrical Flow Matching that enables efficient processing of high-resolution medical images; and (c)~demonstrate that MedSymmFlow delivers calibrated uncertainty estimates natively, without the need for additional post-hoc methods. Through comprehensive experiments across multiple medical imaging benchmarks, we show that MedSymmFlow achieves competitive or superior performance to standard discriminative models, while offering generative and uncertainty estimation capabilities.

\section{Related Work}

\subsection{Flow Matching}

Flow Matching~(FM)~\cite{lipman2022flow} is a generative modeling approach that learns a continuous velocity field to transform a source distribution into a target one. The field is parameterized by a neural network and integrated via numerical solvers, making FM a form of Neural ODE~\cite{chen2018neural}. Unlike Continuous Normalizing Flows~\cite{grathwohl2018ffjord}, FM does not require simulating trajectories during training, improving computational efficiency. It generalizes several generative paradigms, including diffusion models, which can be viewed as a special case with stochastic probability paths.

\begin{figure}[!t]
    \centering
    \includegraphics[width=0.97\linewidth]{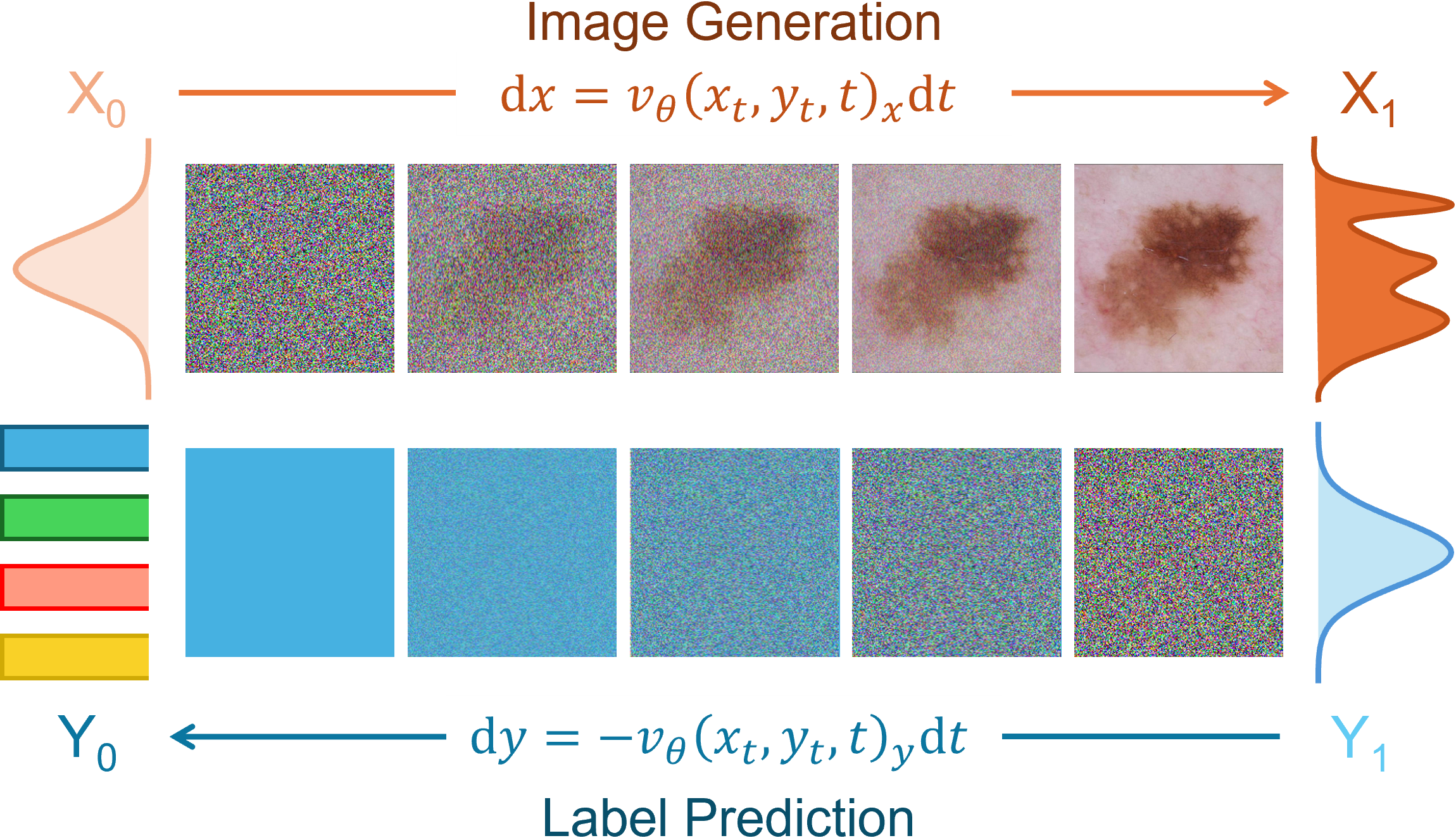}
    \caption{Overview of the MedSymmFlow architecture. The \textcolor{orange}{top row} illustrates the \textcolor{orange}{image generation process}, where samples evolve from Gaussian noise $ x_0 $ to a realistic image $ x_1 $ via forward integration of the conditional vector field. The \textcolor{cyan}{bottom row} shows the \textcolor{cyan}{label prediction process}, in which semantic class labels $ y_1 $ are transformed into predictive estimates $ y_0 $ through reverse-time integration.}
    \label{fig:medsymmflow}
\end{figure}

\subsection{Diffusion Models for Medical Imaging}

Diffusion models have emerged as a powerful class of generative frameworks and are increasingly being adopted across a range of tasks in medical imaging. For data augmentation, they offer a principled approach to synthesizing realistic and diverse samples that capture the variability inherent in clinical data, thus enhancing model robustness and generalization~\cite{guo2024maisi,akrout2023diffusion}. Anatomy-aware augmentation strategies have further expanded this potential by generating contextually relevant variations for tasks such as surgical scene segmentation~\cite{venkatesh2025data}. In segmentation, diffusion-based architectures have demonstrated state-of-the-art performance by leveraging their ability to model complex spatial structures and produce smooth, anatomically accurate masks~\cite{wu2024medsegdiff,wu2024medsegdiffv2}. For anomaly detection, diffusion models are particularly well suited due to their ability to learn high-fidelity distributions of healthy data, which makes semantic deviations, such as tumors or lesions, detectable as out-of-distribution instances~\cite{wyatt2022anoddpm,graham2023unsupervised}. Lastly, in probabilistic classification, diffusion processes have been used to capture predictive uncertainty in a more structured manner than conventional softmax-based classifiers~\cite{favero2025conditional}, which is especially critical in clinical decision making.

\subsection{Generative Classifiers}

Several foundational studies~\cite{hinton2007recognize,ranzato2011deep} have highlighted the critical role of modeling the data distribution to strengthen discriminative feature learning. Early methods, such as deep belief networks~\cite{hinton2006fast}, focused on encoding images into latent spaces that could subsequently support recognition tasks. More recent progress in generative modeling has further demonstrated the effectiveness of compact, transferable representations for global prediction tasks~\cite{he2022masked,croce2020gan}. Furthermore, generative approaches have been shown to improve adversarial robustness and model calibration~\cite{huang2020neural}. Despite these advances, most prior efforts rely either on jointly training generative and discriminative objectives or on fine-tuning generative encoders for specific downstream tasks. Diffusion classifiers~\cite{li2023your,favero2025conditional} have been more recently explored, using diffusion models directly for image classification, although they suffer from high computational demands. SymmFlow~\cite{caetano2025symmetrical} sought to reduce these costs by merging generative and classification processes.

\section{MedSymmFlow}

Symmetrical Flow Matching~(SymmFlow) offers a unified framework for semantic synthesis and classification by modeling opposing flows between images $X$ and semantic content $Y$. It learns a joint velocity field $v_\theta(x_t, y_t, t)$ that simultaneously transports the image $x$ from noise and the semantic target $y$ into noise in the forward direction and reverses this transformation in the backward direction. The core insight is to model generation and prediction as integration through time. Specifically, given a noisy semantic state $y_1$, with $x_t$ and $y_t$ being intermediate states at time $t$, the model predicts the semantic output $y_0$ as
\begin{equation}
    y_0 = y_1 + \int_1^0 v_\theta(x_t, y_t, t)_y \, \text{d}t.
\end{equation}
We propose key modifications to the original SymmFlow framework to enhance its ability to classify and sample data, as illustrated in Figure~\ref{fig:medsymmflow}, and as covered in Subsections~\ref{subsec:rgbmasks} and~\ref{subsec:latentvae}.

\subsection{Semantic Conditioning via RGB Masks}
\label{subsec:rgbmasks}

To enhance semantic conditioning and enable multi-class tasks within a consistent representation, we adopt an RGB encoding scheme where each class label is assigned a unique RGB color code. Instead of discrete categorical or grayscale representations, classification labels are mapped to their corresponding RGB vectors during training and perturbed with uniform noise of amplitude $\beta$. MedSymmFlow~(MSF) maps these continuous RGB signals across the flow.

During inference, the predicted semantic output $\hat{y}_0$ lies in a continuous RGB space. To obtain a final class prediction, we compute the Euclidean distance between $\hat{y}_0$ and each predefined RGB class code $\text{RGB}(c)$, selecting the class with the smallest distance
\begin{equation}
\label{eq:distance}
    \hat{c} = \arg\min_{c \in \mathcal{C}} \| \hat{y}_0 - \text{RGB}(c) \|_2,
\end{equation}
where $\mathcal{C}$ is the set of all class labels. The corresponding distance itself serves as a proxy for uncertainty, 
with larger values indicating higher prediction uncertainty. This approach provides both hard classification decisions and uncertainty estimates using a single generative pass.

\subsection{Latent-Space Implementation}
\label{subsec:latentvae}

To scale MedSymmFlow to higher-resolution~(specifically 224$\times$224~px) medical images, we employ the Variational Autoencoder~(VAE) from the Stable Diffusion architecture, available on HuggingFace\footnote{\url{https://huggingface.co/stabilityai/sd-vae-ft-mse}}. Both the input image $x$ and its corresponding RGB mask $y$ are encoded into latent representations $z_x$ and $z_y$, respectively, using the VAE encoder. The Latent MedSymmFlow~(LatMSF) model is then trained on these representations, facilitating efficient modeling in a reduced-dimensional space. After processing, the predicted semantic output $\hat{z}_y$ is decoded back into RGB space using the VAE decoder.

\section{Methodology}

\subsection{Datasets}

The evaluation is carried out on four MedMNISTv2~\cite{yang2023medmnist} datasets: PneumoniaMNIST, BloodMNIST, DermaMNIST, and RetinaMNIST. These datasets represent diverse imaging modalities and diagnostic tasks, as shown in Table~\ref{tab:datasets}. The datasets are evaluated in low ($28{\times}28$) and high ($224{\times}224$) resolution settings.

\begin{table}[t]
\centering
\caption{Overview of the MedMNIST datasets used in this study.}
\setlength{\tabcolsep}{4pt}
\label{tab:datasets}
\resizebox{\textwidth}{!}{%
\begin{tabular}{l|ccc}
\toprule
\textbf{Dataset} & \textbf{\#Classes} & \textbf{Modality} & \textbf{\#Images~(Train/Val/Test)} \\
\midrule
PneumoniaMNIST & 2 & Chest X-ray & 4,708/524/624 \\
BloodMNIST & 8 & Blood Cell Microscopy & 11,959/1,712/3,421 \\
DermaMNIST & 7 & Dermatoscopy & 7,007/1,003/2,005 \\
RetinaMNIST & 5 & Fundus Camera & 1,080/120/400 \\
\bottomrule
\end{tabular}}
\end{table}

\subsection{Selected Models}

We use the baseline models provided in the MedMNIST benchmark, including ResNet-18, ResNet-50~\cite{he2016deep}, AutoKeras~\cite{jin2019auto}, and Auto-Sklearn~\cite{feurer2019automated}. Furthermore, we evaluate MedViT-S~\cite{manzari2023medvit}, which consistently outperformed the larger variant in the selected datasets. SymmFlow and MSF are also included, with implementation details and computational requirements available in the Appendix.

\subsection{Metrics}

Performance is measured using Accuracy and Area Under the Receiver Operating Characteristic Curve~(AUC), following the MedMNIST evaluation protocol. For SymmFlow and MedSymmFlow, the results are reported as mean and a confidence interval of two standard deviations measured over five independent inference runs, capturing the variation due to probabilistic components.

\subsection{Uncertainty Quantification}

Uncertainty quantification in medical imaging is typically validated by confirming that confident predictions are more likely to be correct, and incorrect predictions are associated with high uncertainty~\cite{favero2025conditional}. To evaluate this property, predictions are ranked by uncertainty, and the most uncertain samples are progressively filtered out, generating an Accuracy-Rejection Curve~(ARC)~\cite{nadeem2009accuracy,barandas2022uncertainty}. For MedSymmFlow, as defined in Equation~\ref{eq:distance}, we use the $\ell_2$-norm distance between the predicted embedding and the prototype of the predicted class as a proxy for uncertainty: larger distances indicate lower confidence. 

\section{Results \& Discussion}

\subsection{Quantitative Results}

Table~\ref{tab:results} presents the classification results in the four MedMNIST datasets. At 28$\times$28 resolution, the original SymmFlow framework with grayscale class masks performs adequately on PneumoniaMNIST and BloodMNIST but drops significantly in performance on DermaMNIST and RetinaMNIST compared to other methods. This performance drop is particularly evident in the AUC scores and stems from a fundamental limitation: encoding class information along a single intensity axis compresses the distance relationships between classes into one dimension, making it difficult for the model to accurately distinguish between multiple semantic categories.

\begin{table*}[!t]
\setlength{\tabcolsep}{4pt}
\centering
\caption{Performance comparison (AUC and Accuracy) across four MedMNIST datasets. ResNet baselines and AutoML systems are compared to SymmFlow and to the MedSymmFlow variants.}
\label{tab:results}
\resizebox{\textwidth}{!}{%
\begin{tabular}{l|ccccccccc}
\toprule
\multirow{2}{*}{\textbf{Model}} & \multicolumn{2}{c}{\textbf{PneumMNIST}} & \multicolumn{2}{c}{\textbf{BloodMNIST}} & \multicolumn{2}{c}{\textbf{DermaMNIST}} & \multicolumn{2}{c}{\textbf{RetinaMNIST}} \\
\cmidrule{2-3} \cmidrule{4-5} \cmidrule{6-7} \cmidrule{8-9}
 & \textbf{AUC} & \textbf{ACC} & \textbf{AUC} & \textbf{ACC} & \textbf{AUC} & \textbf{ACC} & \textbf{AUC} & \textbf{ACC} \\
\midrule
ResNet-18~(28) & 94.4 & 85.4 & \textbf{99.8} & 95.8 & 91.7 & 73.5 & 71.7 & 52.4 \\
ResNet-50~(28) & 94.8 & 85.4 & \underline{99.7} & 95.6 & 91.3 & 73.5 & 72.6 & 52.8 \\
auto-sklearn~(28) & 94.2 & 85.5 & 98.4 & 87.8 & 90.2 & 71.9 & 69.0 & 51.5 \\
AutoKeras~(28) & 94.7 & 87.8 & \textbf{99.8} & 96.1 & 91.5 & 74.9 & 71.9 & 50.3 \\
\midrule
ResNet-18~(224) & 95.6 & 86.4 & \textbf{99.8} & 96.3 & 92.0 & 75.4 & 71.0 & 49.3 \\
ResNet-50~(224) & \underline{96.2} & 88.4 & \underline{99.7} & 95.0 & 91.2 & 73.1 & 71.6 & 51.1 \\
MedViT-S~(224) & \textbf{99.5} & \textbf{96.1} & \underline{99.7} & 95.1 & \textbf{93.7} & 78.0 & \underline{77.3} & \textbf{56.1} \\
\midrule
SymmFlow~(28) & 91.6$\pm$0.8 & \underline{89.4}$\pm$0.5 & 99.1$\pm$0.1 & 96.3$\pm$0.2 & 83.4$\pm$0.6 & 69.3$\pm$0.5 & 70.2$\pm$1.1 & 50.7$\pm$0.7 \\
MSF~(Ours)~(28) & 95.2$\pm$0.4 & 88.0$\pm$0.6 & 99.4$\pm$0.1 & \underline{97.9}$\pm$0.2 & 89.6$\pm$0.5 & \underline{78.8}$\pm$0.8 & 73.1$\pm$0.4 & 51.4$\pm$2.1 \\
LatMSF~(Ours)~(224) & 94.4$\pm$1.7 & \underline{89.4}$\pm$1.1 & \textbf{99.8}$\pm$0.0 & \textbf{99.0}$\pm$0.1 & \underline{92.5}$\pm$0.4 & \textbf{81.0}$\pm$0.6 & \textbf{78.8}$\pm$0.4 & \underline{54.0}$\pm$0.9 \\
\bottomrule
\end{tabular}
}
\end{table*}

Introducing RGB mask conditioning in MedSymmFlow directly addresses this bottleneck. By defining class-specific embeddings along separate RGB channels, the model gains a richer and more structured way to represent distances between classes in the conditioning space. This leads to consistent improvements in AUC, with significant gains in DermaMNIST. Improvements in AUC are mirrored by corresponding increases in classification accuracy, confirming that more expressive conditioning also leads to better decision boundaries.

Extending the RGB-conditioned MSF to a latent model (LatMSF) further amplifies these benefits. The resulting model consistently matches or outperforms ViT-based approaches, trading top positions across tasks, and surpasses traditional CNN classifiers and AutoML systems in overall performance, with the exception of PneumoniaMNIST. In this case, LatMSF shows a noticeable performance drop, likely due to the fixed VAE encoder producing latent codes that omit key discriminative details. As shown in Table~\ref{tab:comp}, MSF models require longer training times and slower inference compared to purely discriminative baselines. This is expected given their generative formulation. In particular, LatMSF incurs additional memory and latency overhead due to the inclusion of the VAE, which contributes to both sampling complexity and model size.

\subsection{Uncertainty Quantification}

\begin{figure*}[!t]
    \centering
    \begin{subfigure}[t]{0.48\textwidth}
        \centering
        \includegraphics[width=\linewidth]{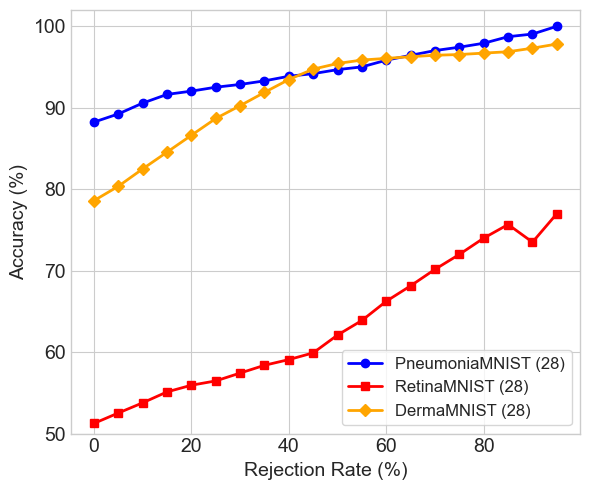}
        \caption{MSF trained on 28$\times$28 images.}
    \end{subfigure}%
    \hfill
    \begin{subfigure}[t]{0.48\textwidth}
        \centering
        \includegraphics[width=\linewidth]{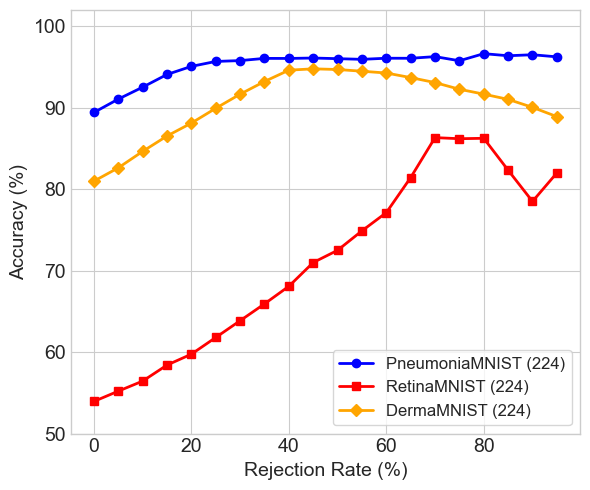}
        \caption{LatMSF trained on 224$\times$224 images.}
    \end{subfigure}
    \caption{Filtering uncertain predictions improves performance, indicating the model is well-calibrated in its confidence. However, excessive filtering eventually removes confident, correct predictions too, leading to diminished results.}
    \label{fig:uncertainty-filtering}
\end{figure*}

Figure~\ref{fig:uncertainty-filtering} shows that for both low- and high-resolution, confidence correlates with accuracy; this indicates that the proposed distance-based uncertainty proxy is well calibrated and can be used to reject low-quality predictions. When the most uncertain predictions are removed, the accuracy increases across all datasets, indicating that the model correctly self-assesses its errors. In high-stakes domains like healthcare, it is safer for models to abstain from uncertain predictions rather than risk confident misclassifications. This trend holds especially well for PneumoniaMNIST and RetinaMNIST. However, at higher resolution, overfiltering begins to degrade DermaMNIST performance, suggesting that confident predictions are also being discarded. For BloodMNIST, which already achieves high accuracy without filtering, performance variations are negligible and omitted.

\subsection{Qualitative Results}

\begin{figure*}[t!]
    \centering
    \begin{subfigure}[t]{0.45\textwidth}
        \centering
        \includegraphics[width=\linewidth]{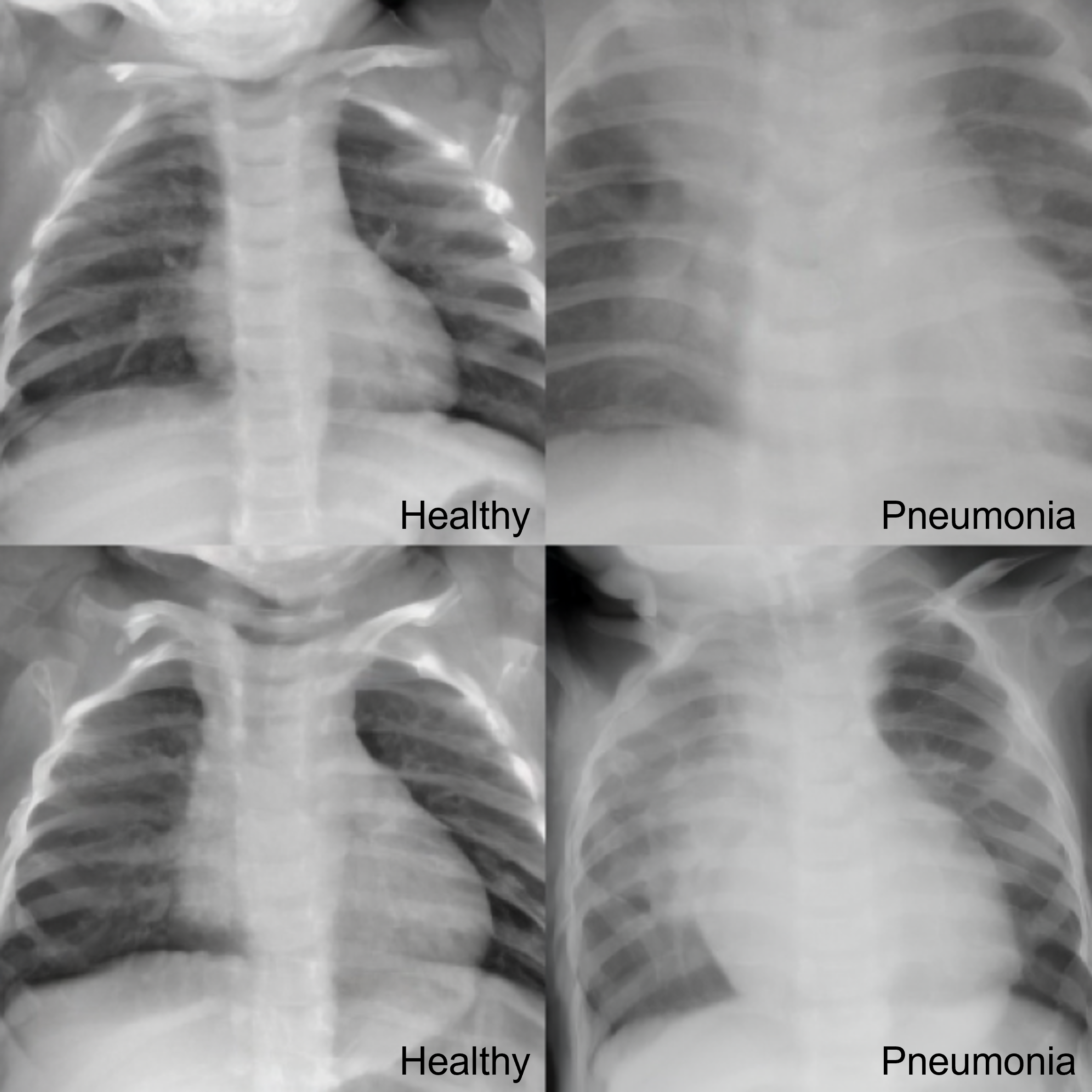}
        \caption{PneumoniaMNIST.}
    \end{subfigure}
    \hfill
    \begin{subfigure}[t]{0.45\textwidth}
        \centering
        \includegraphics[width=\linewidth]{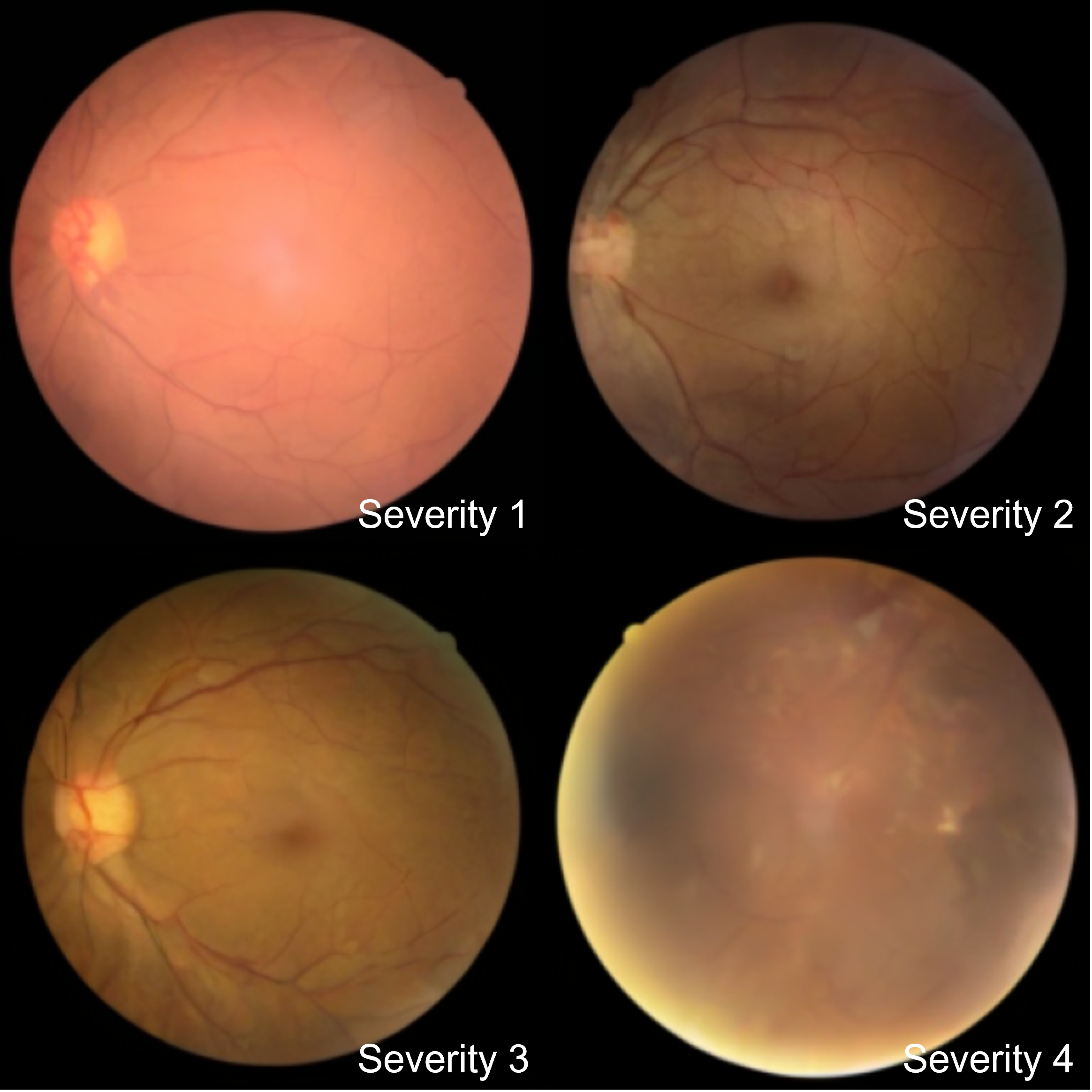}
        \caption{RetinaMNIST.}
    \end{subfigure}
    
    \begin{subfigure}[t]{0.45\textwidth}
        \centering
        \includegraphics[width=\linewidth]{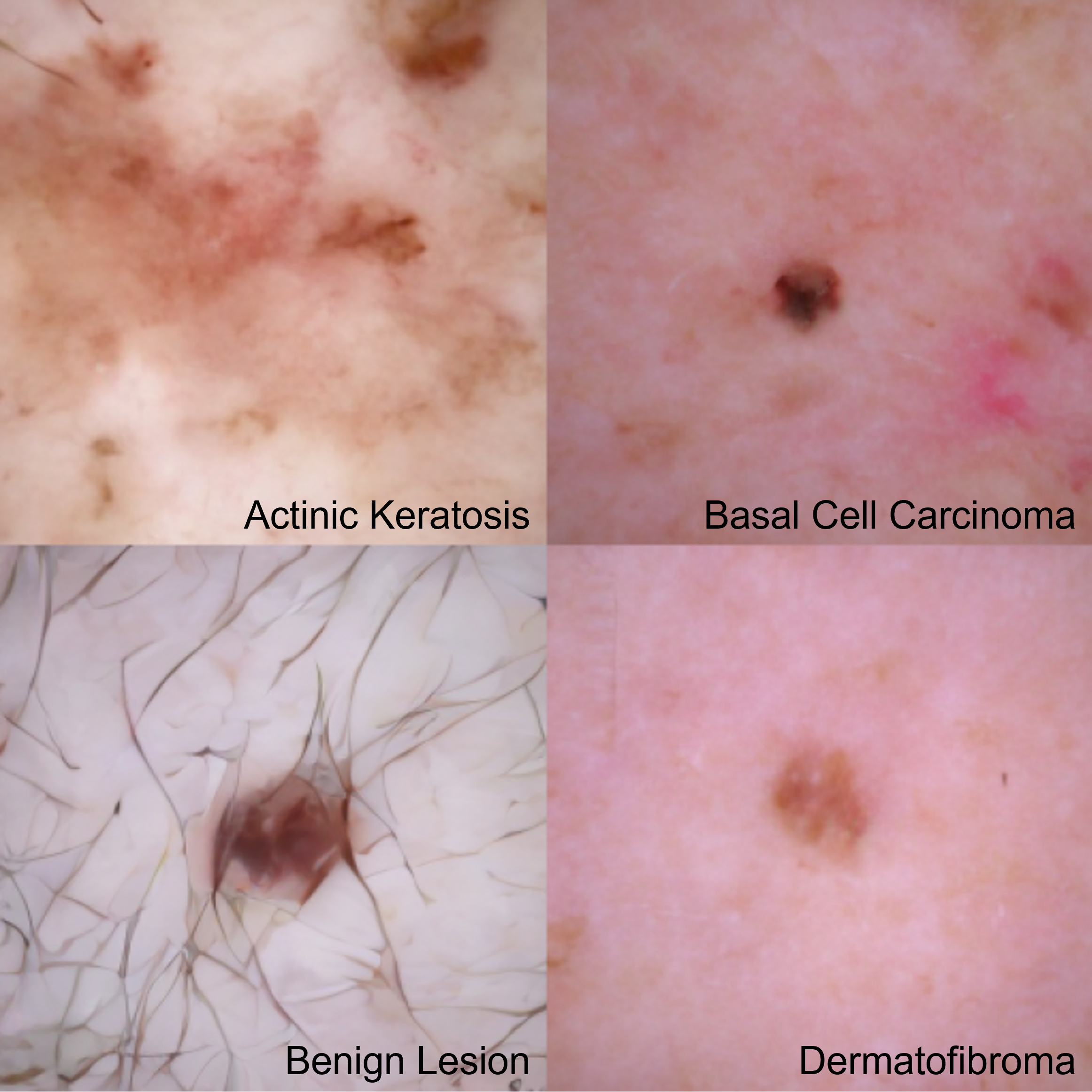}
        \caption{DermaMNIST.}
    \end{subfigure}
    \hfill
    \begin{subfigure}[t]{0.45\textwidth}
        \centering
        \includegraphics[width=\linewidth]{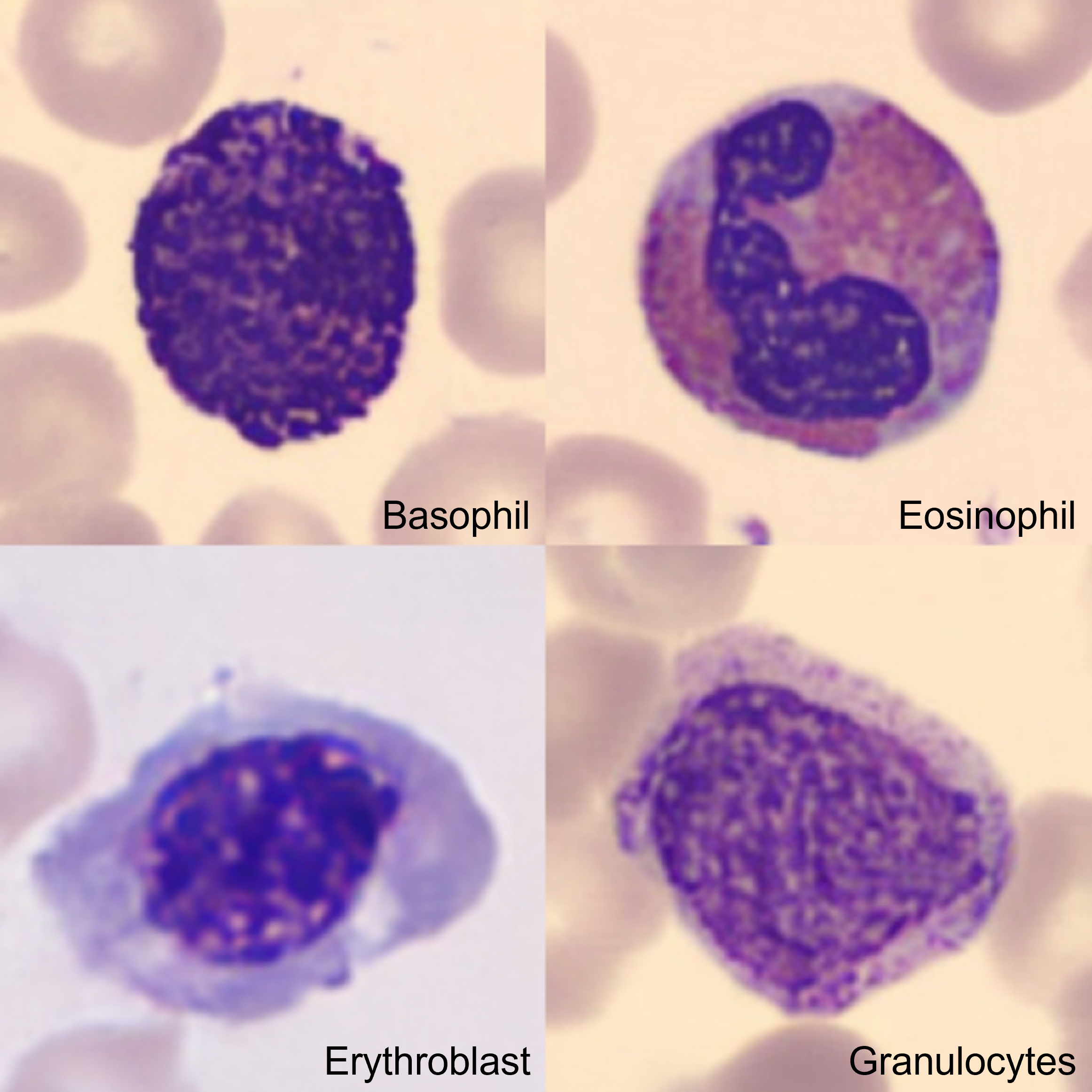}
        \caption{BloodMNIST.}
    \end{subfigure}
    \caption{Samples~(224$\times$224 pixels) generated by the latent models trained on the four selected MedMNIST datasets, using an Euler ODE solver with 25 steps.}
    \label{fig:qualitative-latent}
\end{figure*}

Figure~\ref{fig:qualitative-latent} shows high-resolution images sampled from the latent-space variant of MedSymmFlow. The generated samples exhibit high visual fidelity across datasets, capturing fine-grained details such as skin texture and even hair artifacts, as seen in DermaMNIST, for instance. These results suggest that the model is not only capable of classifying samples, but also learns the underlying semantics of each dataset well enough to generate diverse and realistic samples, reflecting a strong grasp of both anatomical structure and pathological variation.

\section{Limitations \& Future Work}

While MedSymmFlow shows promising results in multi-class classification and image generation, its current evaluation is restricted to curated 2D datasets. This limits the applicability of the model to more realistic clinical scenarios, where most modalities (e.g., MRI, CT) are inherently 3D and present greater anatomical complexity. Extending the model to handle 3D data is therefore a key direction for future work. In addition, more thorough evaluation of more complex and unbalanced data is needed, such as surgical RGB images. Moreover, the inference speed could be improved through model distillation to support deployment in latency-sensitive applications. Finally, we also intend to explore a wider spectrum of tasks, ranging from uncertainty‑aware segmentation to semantically detailed data augmentation, to fully validate the flexibility of the architecture.

\section{Conclusion}

In this work, we present MedSymmFlow, a conditional generative model for semantic image synthesis and uncertainty-aware classification in medical imaging. We introduced a novel conditional formulation based on symmetrical optimal transport with a latent-space formulation, improving joint generation and prediction with uncertainty. Our approach achieves competitive classification accuracy while providing high-fidelity image synthesis, interpretable uncertainty estimates, and the ability to reject low-confidence predictions, essential for safe and reliable deployment in clinical settings.

%
%

\begin{credits}
\subsubsection{\ackname} This research was funded by the Xecs Eureka TASTI Project.
\subsubsection{\discintname}
The authors have no competing interests to declare that are relevant to the content of this article.
\end{credits}
%
%
%
\bibliographystyle{splncs04}

\newpage
\setcounter{page}{1}
\appendix
\section{Supplementary Material}
\subsection{SymmFlow}
\label{ap:sf}

The SymmFlow models were trained on a workstation equipped with an NVIDIA RTX 2080 Ti GPU (11GB VRAM), an Intel Xeon Silver 4216 CPU (2.10 GHz), and 192GB of RAM. The hyperparameters listed in Table~\ref{tab:hyperparams_sf} reflect the exact configuration used during training.

\begin{table}[!ht]
    \centering
    \caption{Hyperparameters used for training each SymmFlow model.}
    \label{tab:hyperparams_sf}
    {\resizebox{0.9\linewidth}{!}{
    \begin{tabular}{@{\extracolsep{6pt}}l|cccc@{}}
        \toprule
        \textbf{Hyperparameter} & \multicolumn{1}{c}{\textbf{PneumMNIST}} & \multicolumn{1}{c}{\textbf{BloodMNIST}} & \multicolumn{1}{c}{\textbf{DermaMNIST}} & \multicolumn{1}{c}{\textbf{RetinaMNIST}} \\
        \midrule
        Channels & 64 & 64 & 64 & 64 \\
        Depth & 2 & 2 & 2 & 2 \\
        Channels Multiple & 1,2,2,2 & 1,2,2,2 & 1,2,2,2 & 1,2,2,2 \\
        Heads & 4 & 4 & 4 & 4 \\
        Head Channels & 64 & 64 & 64 & 64 \\
        Attention Resolution & 2 & 2 & 2 & 2 \\
        Dropout & 0.0 & 0.0 & 0.0 & 0.0 \\
        Batch Size & 128 & 256 & 256 & 256 \\
        GPUs & 1 & 1 & 1 & 1 \\
        Epochs & 1000 & 1000 & 1000 & 2000 \\
        Learning Rate & 5e-4 & 5e-4 & 3e-4 & 5e-4 \\
        Learning Rate Scheduler & Cosine Annealing & Cosine Annealing & Cosine Annealing & Cosine Annealing \\
        Warmup Epochs & 100 & 100 & 100 & 200 \\
        $\beta$ & 1 & 4 & 4 & 3 \\
        \bottomrule
    \end{tabular}}}
\end{table}

\subsection{MedSymmFlow}

The low-resolution MedSymmFlow models were trained under the same workstation setup as the SymmFlow models, with full details provided in Appendix~\ref{ap:sf}. The corresponding hyperparameters are listed in Table~\ref{tab:hyperparams_msf}.

\begin{table}[!ht]
    \centering
    \caption{Hyperparameters used for training each MedSymmFlow model.}
    \label{tab:hyperparams_msf}
    {\resizebox{0.9\linewidth}{!}{
    \begin{tabular}{@{\extracolsep{6pt}}l|cccc@{}}
        \toprule
        \textbf{Hyperparameter} & \multicolumn{1}{c}{\textbf{PneumMNIST}} & \multicolumn{1}{c}{\textbf{BloodMNIST}} & \multicolumn{1}{c}{\textbf{DermaMNIST}} & \multicolumn{1}{c}{\textbf{RetinaMNIST}} \\
        \midrule
        Channels & 64 & 64 & 64 & 64 \\
        Depth & 2 & 2 & 2 & 2 \\
        Channels Multiple & 1,2,2,2 & 1,2,2,2 & 1,2,2,2 & 1,2,2,2 \\
        Heads & 4 & 4 & 4 & 4 \\
        Head Channels & 64 & 64 & 64 & 64 \\
        Attention Resolution & 2 & 2 & 2 & 2 \\
        Dropout & 0.0 & 0.0 & 0.0 & 0.0 \\
        Batch Size & 128 & 256 & 256 & 256 \\
        GPUs & 1 & 1 & 1 & 1 \\
        Epochs & 1000 & 1000 & 1000 & 2000 \\
        Learning Rate & 5e-4 & 5e-4 & 3e-4 & 5e-4 \\
        Learning Rate Scheduler & Cosine Annealing & Cosine Annealing & Cosine Annealing & Cosine Annealing \\
        Warmup Epochs & 100 & 100 & 100 & 200 \\
        $\beta$ & 4 & 4 & 4 & 4 \\
        \bottomrule
    \end{tabular}}}
\end{table}

The latent MedSymmFlow models were trained on a workstation featuring an NVIDIA RTX 3090 Ti GPU (24GB VRAM), an Intel Xeon Silver 4216 CPU (2.10 GHz), and 192GB of RAM. The training setup and hyperparameters used are detailed in Table~\ref{tab:hyperparams_latmsf}.

\begin{table}[!ht]
    \centering
    \caption{Hyperparameters used for training each Latent MedSymmFlow model.}
    \label{tab:hyperparams_latmsf}
    {\resizebox{0.9\linewidth}{!}{
    \begin{tabular}{@{\extracolsep{6pt}}l|cccc@{}}
        \toprule
        \textbf{Hyperparameter} & \multicolumn{1}{c}{\textbf{PneumMNIST}} & \multicolumn{1}{c}{\textbf{BloodMNIST}} & \multicolumn{1}{c}{\textbf{DermaMNIST}} & \multicolumn{1}{c}{\textbf{RetinaMNIST}} \\
        \midrule
        Channels & 128 & 128 & 128 & 128 \\
        Depth & 2 & 2 & 2 & 2 \\
        Channels Multiple & 1,2,2,2 & 1,2,2,2 & 1,2,2,2 & 1,2,2,2 \\
        Heads & 4 & 4 & 4 & 4 \\
        Head Channels & 64 & 64 & 64 & 64 \\
        Attention Resolution & 2 & 2 & 2 & 2 \\
        Dropout & 0.0 & 0.0 & 0.0 & 0.0 \\
        Batch Size & 64 & 64 & 128 & 128 \\
        GPUs & 1 & 1 & 1 & 1 \\
        Epochs & 2000 & 1000 & 1000 & 2000 \\
        Learning Rate & 3e-4 & 4e-4 & 5e-4 & 5e-4 \\
        Learning Rate Scheduler & Cosine Annealing & Cosine Annealing & Cosine Annealing & Cosine Annealing \\
        Warmup Epochs & 200 & 100 & 100 & 200 \\
        $\beta$ & 4 & 4 & 4 & 4 \\
        \bottomrule
    \end{tabular}}}
\end{table}

\subsection{Computational Requirements}

To ensure a fair comparison, all models were retrained and evaluated on the workstation equipped with an NVIDIA RTX 3090 Ti GPU (24GB VRAM), an Intel Xeon Silver 4216 CPU (2.10 GHz), and 192GB of RAM. For consistency, BloodMNIST was used as a benchmark, the largest of the MedMNISTv2 datasets selected. Training time is reported as the total time required to reach convergence. Inference latency and memory usage were measured in 100 single-image batches; we report the average latency and maximum GPU memory allocated as measured by \texttt{torch.cuda.max\_memory\_allocated}. The results are summarized in Table~\ref{tab:comp}.

\begin{table}[!ht]
    \centering
    \caption{Computational requirements for training and running inference on the evaluated models.}
    \label{tab:comp}
    {\resizebox{\linewidth}{!}{
    \begin{tabular}{@{\extracolsep{6pt}}l|cccc@{}}
    \toprule
      & \textbf{Training} & \multicolumn{3}{c}{\textbf{Inference}} \\
     \cline{3-5}
     \textbf{Model} & \textbf{Time~(min)} &  \textbf{Memory~(MB)} & \textbf{Classification Latency~(ms)} & \textbf{Sampling Latency~(ms)}\\
     \midrule
     ResNet-18~(28) & 12 & 63.6 & 0.4 & ---\\
     ResNet-18~(224) & 99 & 118.2 & 3.4 & ---\\
     ResNet-50~(28) & 45 & 113.4 & 0.5 & ---\\
     ResNet-50~(224) & 372 & 286.4 & 14.5 & ---\\
     MedViT-S~(224) & 395 & 170.7 & 29.0 & --- \\
     \midrule
     SymmFlow~(28) & 178 & 68.3 & 15.3 & 360.1 \\
     MSF~(28) & 180 & 68.3 & 15.3 & 360.5 \\
     LatMSF~(224) & 2,302 & 712.1 & 68.2 & 540.1 \\
     \bottomrule
    \end{tabular}}}
\end{table}
\end{document}